\documentclass[10pt,twocolumn,letterpaper]{article}

\usepackage{cvpr}
\usepackage{times}
\usepackage{epsfig}
\usepackage{graphicx}
\usepackage{amsmath}
\usepackage{amssymb}
\usepackage{subfigure}
\usepackage{multirow}
\usepackage{pdfpages}

\usepackage[breaklinks=true,bookmarks=false]{hyperref}

\cvprfinalcopy 

\begin{document}
%
\title{Responses to Critiques on Machine Learning of Criminality Perceptions \\
(Addendum of arXiv:1611.04135)}

\author{Xiaolin Wu \\
McMaster University \\
Shanghai Jiao Tong University \\
{\tt\small xwu510@gmail.com}
\and
Xi Zhang\\
Shanghai Jiao Tong University\\
{\tt\small zhangxi\_19930818@sjtu.edu.cn}
}

\maketitle


In November 2016 we submitted to arXiv our paper ``Automated Inference on Criminality Using Face Images''. It generated a great deal of discussions in the Internet and some media outlets.  Our work is only intended for pure academic discussions; how it has become a media consumption is a total surprise to us.  

Although in agreement with our critics on the need and importance of policing AI research for the general good of the society, we are deeply baffled by the ways some of them mispresented our work, in particular the motive and objective of our research.

\section{Name calling}

It should be abundantly clear, for anyone who reads our paper with a neutral mind setting, that our only motive is to know if machine learning has the potential of acquiring humanlike social perceptions of faces, despite the complexity and subtlety of such perceptions that are functions of both the observed and the observer.  Our inquiry is to push the envelope and extend the research on automated face recognition from the biometric dimension (e.g., determining the race, gender, age, facial expression, etc.) to the sociopsychological dimension.  We are merely interested in the distinct possibility of teaching machines to pass the Turing test on the task of duplicating humans in their first impressions (e.g., personality traits, mannerism, demeanor, etc.) of a stranger.  The face perception of criminality was expediently (unfortunately to us in hindsight) chosen as an easy test case, at least in our intuition as explained in our paper:

\begin{quote}
\emph{
``For validating the hypothesis on the correlations between the innate traits and social behaviors of a person and the physical characteristics of that person’s face, it would be hard pushed to find a more convincing experiment than examining the success rates of discriminating between criminals and non-criminals with modern automatic classifiers. These two populations should be among the easiest to differentiate, if social attributes and facial features are correlated, because being a criminal requires a host of abnormal (outlier) personal traits. If the classification rate turns out low, then the validity of face-induced social inference can be safely negated.''
}
\end{quote}

By a magical stretch of imagination, few of our critics intertwine the above passage into some of our honest observations and morph them into the following deduction of, they insist, ours:

\begin{quote}
\emph{
``Those with more curved upper lips and eyes closer together are of a lower social order, prone to (as Wu and Zhang put it) ``a host of abnormal (outlier) personal traits'' ultimately leading to a legal diagnosis of ``criminality'' with high probability.''
}
\end{quote}

We agree that the pungent word criminality should be put in quotation marks; a caveat about the possible biases in the input data should be issued.  Taking a court conviction at its face value, i.e., as the ``ground truth'' for machine learning, was indeed a serious oversight on our part.  However, throughout our paper we maintain a sober neutrality on whatever we might find; in the introduction, we declare

\begin{quote}
\emph{
``In this paper we intend not to nor are we qualified to discuss or debate on societal stereotypes, rather we want to satisfy our curiosity in the accuracy of fully automated inference on criminality. At the onset of this study our gut feeling is that modern tools of machine learning and computer vision will refute the validity of physiognomy, although the outcomes turn out otherwise.''
}
\end{quote}

Nowhere in our paper advocated the use of our method as a tool of law enforcement, nor did our discussions advance from correlation to causality.  But still we got interpreted copiously by some with an insinuation of racism.  This is not the way of academic exchanges we are used to.

Now we came to regret our choice of the terminology ``physiognomy'', the closest English translation for the Chinese folklore term ``Mian Xiang Xue''.
We were not sensitive enough to the inherent dirty connotation of the word in the English speaking academia; merely using the term deserves the label of ``scientific racism''?

\section{Base Rate Fallacy}

While some of our critics proclaimed \emph{``writing for a wide audience: not only for researchers $\cdots$''}, they conveniently did not address the clear symptom of ``base rate fallacy'' exhibited by non-technical types in internet blogs and some media coverage.  Many reports and comments on our research overemphasize on high success rates (granted, still in need of more rigorous validation) of our classifiers; they leap from these numbers to the ``grave'' danger of AI. Sorry, we have to bore technical readers by pointing out a trap of invalid reasoning called the base rate fallacy: the mind tends to lock on a high specific probability (the 89\% true positive rate of our CNN classifier) and ignore the very low underlying background probability (0.36\% crime rate in China).

If Wu is tested positive by our ``criminality'' classifier, how high a chance will he break the law?   Nine of ten odds as cried out by a journalist?  By Bayesian statistical inference (a rudimentary knowledge to research communities) Wu's chance of committing a crime is
\begin{equation}
\centering
P(C|+) = \frac{P(+|C)P(C)} { P(+|C)P(C)+P(+|N)(1-P(C))}
\end{equation}
where $P(+|C)=0.89$ is the probability that a convicted Chinese adult male is tested positive by our CNN face classifier,
$P(C)=0.0036$  is the crime rate of China, and $P(+|N)=0.07$ is the probability that a non-criminal Chinese adult male is tested positive.  Plugging all these numbers into the Bayes formula, Wu is found to have a probability of only 4.39\% to break the law, despite being tested positive by a method of unbelievably high accuracy.  Hopefully, this mathematical journey from 89\% to 4.39\% will put many of our critics at ease.  Having done the above exercise we want to stress again our strong opposition against any practical uses of our methods, not only because their accuracies fall far below any minimum standard.

Base rate fallacy is an old trick used by irresponsible media to sensationalize or exaggerate either the virtue or vice of new (unfamiliar/mysterious to general public) technological and scientific advances.  It can be easily manipulated to instill irrational fears or hopes into ordinary folks about the AI research.

\section{Garbage in}

As much irked by the intellectually chauvinistic tone of few of our critics, we do not dispute their progressive social values. There is really no need to parade infamous racists in chronic order with us inserted at the terminal node.  But the objectivity does exist, at least in theory, independent of whatever prevailing social norms.

With a Ph.D in computer science, we know all too well ``garbage in and garbage out''.  However, some of our critics seemed to suggest that machine learning tools cannot be used in social computing simply because no one can prevent the garbage of human biases from creeping in.  We do not share their pessimism.  Like most technologies, machine learning is neutral.  If it can be used to reinforce human biases in social computing problems as some argued, then it can also be used to detect and correct human biases (prejudice).  They worry about the feedback loop but conveniently do not see that the feedback can be either positive or negative. Granted, the criminality is a highly delicate and complex matter; however, well-trained human experts can strive to ensure the objectivity of the training data, i.e., rendering correct legal decisions independent of facial appearance of the accused.   If the labeling of training face images, or any other type of data for that matter, is free of human biases, then the advantages of automated inference over human judgment in objectivity cannot be denied.

Even in the presence of label noises, regardless they are random or systematic, scientific methods do exist to launder and restore/enhance credence to the results of statistical inferences.  Should we forego scientific knowledge for any shade of populism?

\section{Risk of Overfitting}

Our critics are quick to point out the relatively small sample set used in our experiments and the risk of data overfitting. We are sorely aware of this weakness but cannot get more ID images of convicted Chinese males for obvious reasons (the sort of ongoing publicity might have dashed all our hopes to enrich our data set).  However, we did make our best efforts to validate our findings in Section 3.3 of our paper, which opened as follows but completely ignored by our critics.

\begin{quote}
\emph{
``Given the high social sensitivities and repercussions of our topic and skeptics on physiognomy [19], we try to excise maximum caution before publishing our results. In playing devil's advocate, we design and conduct the following experiments to challenge the validity of the tested classifiers $\cdots$''
}
\end{quote}

We randomly label the faces of our training set as negative and positive instances with equal probability, and run all four classifiers to test if any of them can separate the randomly labeled face images with a chance better than flipping a coin.  All face classifiers fail the above test and other similar, more challenging tests (refer to our paper for details).  These empirical findings suggest that the good classification performances reported in our paper are not due to data overfitting; otherwise, given the same size and type of sample set, the classifiers would also be able to separate randomly labeled data.

Although our sample set of 2000 face images is far from being large, it is already one order of magnitude larger than the one used in a similar study carried out by a team of Cornell researchers.  More importantly, the adequacy of the sample set size is a function of the variability of the data on hand.  In our case, all face images are of the same race, gender, and the head pose; all facial landmark points are aligned via an affine transform.  All these factors greatly reduce the data variability and thus the risk of data overfitting.

\section{White Collar}

Regarding to the question of our critics on the wearing of white-collared shirts by some men but not by others in the ID portraits used in our experiments, we did segment the face portion out of all ID images.  The face-only images are used in training and testing.  The complete ID portraits are presented in our paper only for illustration purposes.  We did not spell out this data preparation detail because it is a standard practice of the field.

Nevertheless, the cue of white collar exposes an important detail that we owe the readers an apology. That is, we could not control for socioeconomic status of the gentlemen whose ID photos were used in our experiments.  Not because we did not want to, but we did not have access to the metadata due to confidentiality issues.  Now reflecting on this nuance, we speculate that the performance of our face classifiers would drop if the image data were controlled for socioeconomic status.  Immediately a corollary of social injustice might follow, we suppose.  In fact, this is precisely why we thought that our results could have significance to social sciences.

\section{Smiley}
In our experiments, we did control facial expressions (e.g., smile and sad) but not faint micro-expressions (e.g., relaxed vs. strained).  We intend to exert much tighter control on facial micro-expressions in the future as soon as a reliable algorithm reaches the sophistication to do so.  

\begin{figure}
\centering
\subfigure[]{
\includegraphics[width=0.2\columnwidth]{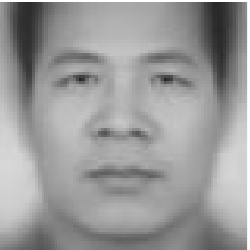}
}
\subfigure[]{
\includegraphics[width=0.2\columnwidth]{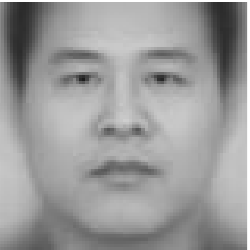}
}
\subfigure[]{
\includegraphics[width=0.2\columnwidth]{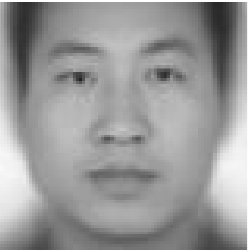}
}
\subfigure[]{
\includegraphics[width=0.2\columnwidth]{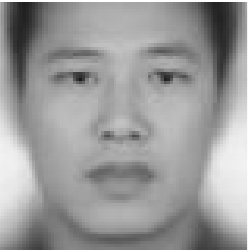}
}
\subfigure[]{
\includegraphics[width=0.2\columnwidth]{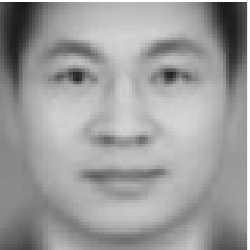}
}
\subfigure[]{
\includegraphics[width=0.2\columnwidth]{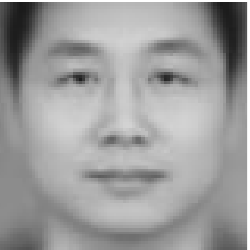}
}
\subfigure[]{
\includegraphics[width=0.2\columnwidth]{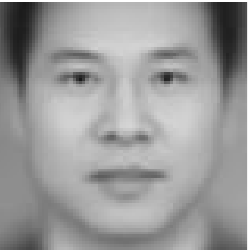}
}
\caption{(a), (b), (c) and (d) are the four face subtypes corresponding to four cluster centroids on the manifold of $S_c$; 
(e), (f) and (g) are the three subtypes of non-criminal faces corresponding to three cluster centroids on the manifold of $S_n$. }
\label{typical}
\end{figure}

The critique that our methods can be reduced to a simple discriminator of smiling versus not-smiling has given us a new angle of scrutiny.  Some Westerns think, by 
staring at the representative faces (``subtypes'') found by our clustering analysis (reproduced as Figure \ref{typical}) of the two populations, that the faces in the bottom row have hints of smile whereas those in the top row do not.
But our Chinese students and colleagues, even after being prompted to consider the cue of smile, fail to detect the same.  Instead, they only find the faces in the bottom row appearing somewhat more relaxed than those in the top row. Perhaps, the different perceptions here are due to cultural differences.

All criminal ID photos are government issued, but not mug shots. To our best knowledge, they are normal government issued ID portraits like those for driver's license in USA.  In contrast, most of the noncriminal ID style photos are taken officially by some organizations (such as real estate companies, law firms, etc.) for their websites.  We stress that they are not selfies.

In our paper, we have also taken steps to prevent the machine learning methods, CNN in particular, from picking up superficial differences between images, such as compression noises and different cameras (Section 3.3).

In conclusion, we appreciate all questions and discussions surrounding our paper, but categorically reject the distortions of our intentions, which are not helpful for the progress and health of the AI research.

\clearpage



\section*{Supplementary material}

# Automated Inference on Criminality using Face Images

Xiaolin Wu
McMaster University
Shanghai Jiao Tong University
xwu510@gmail.com

Xi Zhang
Shanghai Jiao Tong University
zhangxi_19930818@sjtu.edu.cn

## Abstract

*We study, for the first time, automated inference on criminality based solely on still face images, which is free of any biases of subjective judgments of human observers. Via supervised machine learning, we build four classifiers (logistic regression, KNN, SVM, CNN) using facial images of 1856 real persons controlled for race, gender, age and facial expressions, nearly half of whom were convicted criminals, for discriminating between criminals and non-criminals. All four classifiers perform consistently well and empirically establish the validity of automated face-induced inference on criminality, despite the historical controversy surrounding this line of enquiry. Also, some discriminating structural features for predicting criminality have been found by machine learning. Above all, the most important discovery of this research is that criminal and non-criminal face images populate two quite distinctive manifolds. The variation among criminal faces is significantly greater than that of the non-criminal faces. The two manifolds consisting of criminal and non-criminal faces appear to be concentric, with the non-criminal manifold lying in the kernel with a smaller span, exhibiting a law of "normality" for faces of non-criminals. In other words, the faces of general law-biding public have a greater degree of resemblance compared with the faces of criminals, or criminals have a higher degree of dissimilarity in facial appearance than non-criminals.*

## 1. Introduction

Motivated by many commercial applications of artificial intelligence and man-machine interfaces, the research communities of pattern recognition and computer vision have devoted a great deal of efforts to the recognition and manipulation of human faces [11, 31, 40, 35], and achieved measured successes. But very little research has been done on analyzing and quantifying social perception and attributes of faces [33], although this subject is of great importance to many academic disciplines, such as social psychology, management science, criminology, etc.

In all cultures and all periods of recorded human history, people share the belief that the face alone suffices to reveal innate traits of a person. Aristotle in his famous work Prior Analytics asserted, "It is possible to infer character from features, if it is granted that the body and the soul are changed together by the natural affections". Psychologists have known, for as long as a millennium, the human tendency of inferring innate traits and social attributes (e.g., the trustworthiness, dominance) of a person from his/her facial appearance, and a robust consensus of individuals' inferences . These are the facts found through numerous studies [3, 39, 5, 6, 10, 26, 27, 34, 32].

Independent of the validity of pedestrian belief in the (pseudo)science of physiognomy, a tantalizing question naturally arises: what facial features influence average Joes' impulsive and yet consensual judgments on social attributes of a non-acquaintance member of their own specie? Attempting to answer the question, Todorov and Oosterhof proposed a data-driven statistical modeling method to find visual determinants of social attributes by asking human subjects to score four percepts: dominance, attractiveness, trustworthiness, and extroversion, based on first impression of static face images [33]. This method can synthesize a representative (average) face image for a set of input face images scored closely on any of the four aforementioned social percepts. The ranking of these synthesized face images by subjective scores (e.g., from least to most trustworthy looking) apparently agrees with the intuition of most people.

Following the consensus in social perception from facial appearance, arrives the next even bigger speculation: is there any diagnostic merit of the face-induced inferences on an individual's social attributes? In this paper we intend not to nor are we qualified to discuss or debate on societal stereotypes, rather we want to satisfy our curiosity in the accuracy of fully automated inference on criminality. At the onset of this study our gut feeling is that modern tools of machine learning and computer vision will refute the validity of physiognomy, although the outcomes

turn out otherwise. Adopting an approach of supervised machine learning, we build four classifiers (logistic regression [17], $K$-nearest neighbor (KNN) [2], Support Vector Machine (SVM) [14], and Convolutional Neural Networks (CNN) [21]) using facial images of 1856 real persons, half of whom were convicted criminals, and evaluate the performances of these learnt classifiers. The work presented by this paper is the first of its kind to our best knowledge.

As modern machine learning algorithms can match and even exceed human's performance in face recognition [24], it becomes irresistible to pursue automated inference on criminality. Just like in face recognition, first impressions from faces are raw, spontaneous assessments that are a result of perceiving rather than reasoning. Even 3-to-4-year-olds reach a degree of agreement with adults in faced-induced judgment of social attributes [13]. Therefore, we believe that social perception of faces should be a worthy and challenging topic for computer vision and machine learning.

For validating the hypothesis on the correlations between the innate traits and social behaviors of a person and the physical characteristics of that person's face, it would be hard pushed to find a more convincing experiment than examining the success rates of discriminating between criminals and non-criminals with modern automatic classifiers. These two populations should be among the easiest to differentiate, if social attributes and facial features are correlated, because being a criminal requires a host of abnormal (outlier) personal traits. If the classification rate turns out low, then the validity of face-induced social inference can be safely negated.

In contrast to a large body of research literature on the prevalence and consequences of appearance-induced inference on personal traits in the field of psychology, relatively little study has been done on the accuracy of character inference based solely on still face images [37]. This is probably due to, aside from the historical controversies surrounding the inquiry and stigmas associated with social Darwinism, the difficulty to neutralize all possible prejudice and preconditioning of human experimenters and subjects when assessing the accuracy of face-induced inference on socially charged matters such as criminality. In this work, we adopt the approach of data-driven machine learning to fully automate the assessment process, and purposefully take any subtle human factors out of the assessment process.

Unlike a human examiner/judge, a computer vision algorithm or classifier has absolutely no subjective baggages, having no emotions, no biases whatsoever due to past experience, race, religion, political doctrine, gender, age, etc., no mental fatigue, no preconditioning of a bad sleep or meal. The automated inference on criminality eliminates the variable of meta-accuracy (the competence of the human judge/examiner) all together. Besides the advantage of objectivity, sophisticated algorithms based on machine learning may discover very delicate and elusive nuances in facial characteristics and structures that correlate to innate personal traits and yet hide below the cognitive threshold of most untrained nonexperts. This is at least a distinct theoretical possibility.

Although being the first to study automatic face-induced inference on criminality, our findings are significant in that they are derived not only by more advanced data analysis algorithms but also from more realistic, higher quality data set than in previous similar studies carried out by psychologists who use traditional methods. Our data for the training of the classifiers are standard ID photographs of real persons controlled for race, gender, age, and facial expression. In contrast, other studies used sample face images that were synthetic and generated by either 2D or 3D face models based on eigenfaces [35, 8, 9]. However, it is not clear that the synthetic faces, generated by random perturbing parameters of a mathematical face model as done in [35, 8], truly and fairly represent the population. Only in Valla et al.'s study on face-induced inference on criminality, real face images of male Caucasians were used [37]. But in this study humans not computers performed the classification task.

This paper is structured as follows. In Section 2, we detail the preparations of experiment data and the control of variables to set the subsequent data-driven machine learning methods for inferring criminality on solid ground. In Section 3, we represent our methods for assessing the accuracy of automated face-induced inference on criminality and report our findings, which offer strong evidences pointing to high success rate of automatic classification between criminals and non-criminals based on ID photos. The positive results are cross-validated vigorously. In Section 4, we explore discriminating facial features that contribute to the success of automated face-induced inference on criminality. In Section 5, we try to gain some insight in the underlying mechanism for the separability of criminal and non-criminal faces via cluster analysis on manifolds. It is found that criminal face images and non-criminal face images populate two quite distinctive manifolds. The variation among criminal faces is significantly greater than that of the non-criminal faces. The two manifolds formed respectively by the two data sets of criminal and non-criminal faces are concentric, with the manifold for the non-criminal face images lying in the kernel with a smaller span. This newly discovered knowledge suggests a law of normality for faces of non-criminals: Given the race, gender and age, the faces of general law-biding public have a greater degree of resemblance compared with the faces of criminals. In other words, criminals have a significantly higher degree of dissimilarity in facial appearance than normal population. Section 6 concludes the paper.

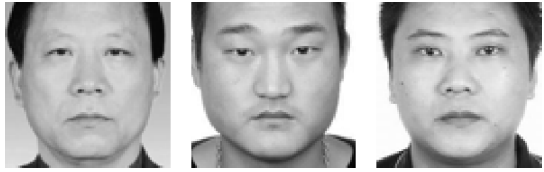

(a) Three samples in criminal ID photo set $S_c$.

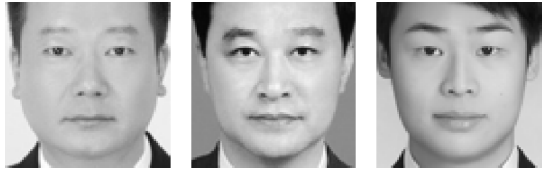

(b) Three samples in non-criminal ID photo set $S_n$

Figure 1. Sample ID photos in our data set.

## 2. Data preparation

In order to conduct our experiments and draw conclusions with strict control of variables, we collected 1856 ID photos that satisfy the following criteria: Chinese, male, between ages of 18 and 55, no facial hair, no facial scars or other markings, and denote this data set by $S$. Set $S$ is divided into two subsets $S_n$ and $S_c$ for non-criminals and criminals, respectively. Subset $S_n$ contains ID photos of 1126 non-criminals that are acquired from Internet using the web spider tool; they are from a wide gamut of professions and social status, including waiters, construction workers, taxi and truck drivers, real estate agents, doctors, lawyers and professors; roughly half of the individuals in subset $S_n$ have university degrees.

Subset $S_c$ contains ID photos of 730 criminals, of which 330 are published as wanted suspects by the ministry of public security of China and by the departments of public security for the provinces of Guangdong, Jiangsu, Liaoning, etc.; the others are provided by a city police department in China under a confidentiality agreement. We stress that the criminal face images in $S_c$ are normal ID photos *not* police mugshots. Out of the 730 criminals 235 committed violent crimes including murder, rape, assault, kidnap and robbery; the remaining 536 are convicted of non-violent crimes, such as theft, fraud, abuse of trust (corruption), forgery and racketeering. Some sample ID photos in $S_c$ and $S_n$ are displayed in Figure 1. The individuals in $S_c$ and $S_n$ are residents of a very large geographical areas, stretching from the northeast all the way to the far south of China and including poor and very rich provinces of the country.

In all selected ID photos, only the region of the face and upper neck is extracted and the background is removed. All the extracted faces are normalized in size and aligned into an $80 \times 80$ image. Although all test face images are ID photos acquired with uniform frontal lighting, we still take extra measures to neutralize any possible effects of varied illumination conditions. Only the luminance component of all color face images is used to factor out the spectrum of the lighting and the skin color. Moreover, all resulting grey scale images are normalized to have the same intensity distribution or the same overall tone production. This is done by matching the histogram of every input image to the average histogram for the entire data set of 1856 grey scale face images.

All ID photos in $S$ are JPEG compressed with QP factor of 90 or higher. Still we applied JPEG soft decoding techniques [23, 28] to remove small (perceptually transparent) compression noises; in the process any device-dependent, signal-level signatures are destroyed as well.

## 3. Validity of Face Classifiers on Criminality

As argued in the introduction, one way of assessing the accuracy of the automated inference on criminality based solely on still face images is to build and test classifiers with modern machine learning techniques. This section presents the design and results of the classification experiments.

### 3.1. Methods

In order to prove or disprove the hypothesis that still face images suffice to distinguish criminals and non-criminals, we try to make our investigations as thorough as possible. We run four different classification methods, logistic regression, KNN, SVM and CNN, on the image data set $S$ prepared as above.

As the first three classification methods work on image features, we run them and evaluate their performances on a wide range of features, including 1. Facial landmark points like eye corners, mouth corners and tip of the nose, etc.; 2. Facial feature vector generated by modular PCA [18]; 3. Facial feature vector based on Local Binary Pattern (LBP) histograms [1]; 4. The concatenation of the above three feature vectors. We stress that the landmark points are defined of strategic positions on a face, hence they are features that are beyond signal level and invariant to source cameras.

Our convolutional neural network is constructed by retraining the parameters of every layer in AlexNet [21] while retaining its architecture.

Define the criminal subset $S_c$ as the positive class and the non-criminal subset $S_n$ as the negative class. We perform 10-fold cross validation for all possible combinations of the three feature-driven classifiers and the four types of feature vectors, plus data-driven CNN without explicit feature vector; altogether thirteen cases (3 classifiers $\times$ 4 feature vectors plus CNN) of 10-fold cross validation type. In the interest of statistic significance we repeated the cross validation for each of the thirteen cases ten times with different random seeds. In each of these (13 cases $\times$ 10 runs)

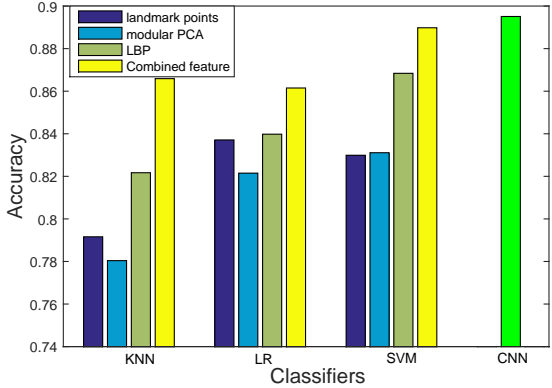

Figure 2. Accuracy of all four classifiers in all thirteen cases.

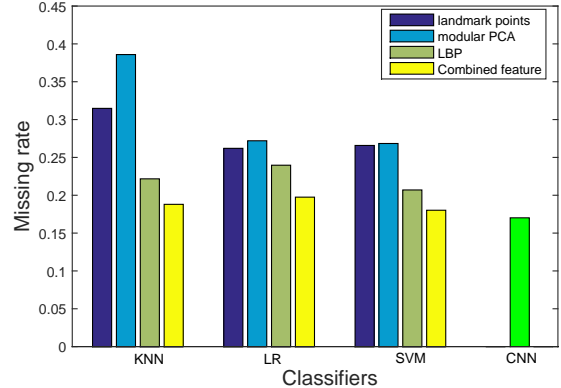

Figure 3. Missing rate of all four classifiers in all thirteen cases.

| Classifiers | CNN | SVM | KNN | LR |
|---|---|---|---|---|
| AUC | 0.9540 | 0.9303 | 0.8838 | 0.8666 |

Table 1. The AUC results for the four tested face classifiers on criminality.

130 experiments, we examine the rate of correctly classifying a member of $S$ into $S_n$ or $S_c$, and then average the rates of each case over ten runs.

### 3.2. Results

In Figure 2, we plot the accuracies of all four classifiers in the above thirteen cases. As expected, the state-of-the-art CNN classifier performs the best, achieving $89.51\%$ accuracy. The relatively high accuracy of CNN is also paralleled by all other three classifiers which are only few percentage points behind in the success rate of classification. These highly consistent results are evidences for the validity of automated face-induced inference on criminality, despite the historical controversy surrounding the topic. We also plot the missing rate and false alarm rate for the four tested classifiers in Figure 3 and Figure 4.

To measure the sensitivities of the automatic, data-driven, binary face classifiers for criminality in relation to their false alarm rates, we plot the ROC curves for the four tested classifiers (see Figure 5), and report the corresponding AUC results in Table 1 as well. By these classification performance metrics, the prediction power of the proposed approach of automated face-induced inference on criminality is established.

### 3.3. Validation

Given the high social sensitivities and repercussions of our topic and skeptics on physiognomy [19], we try to excise maximum caution before publishing our results. In playing devil's advocate, we design and conduct the following experiments to challenge the validity of the tested

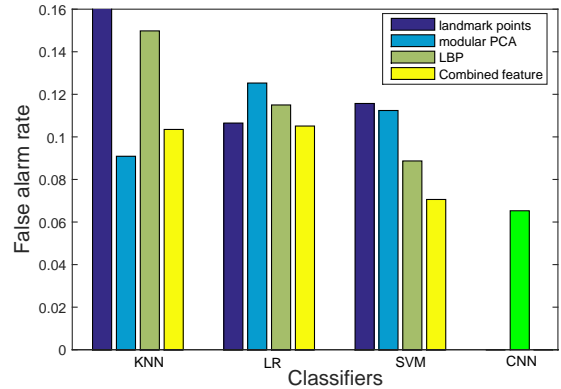

Figure 4. False alarm rate of all four classifiers in all thirteen cases.

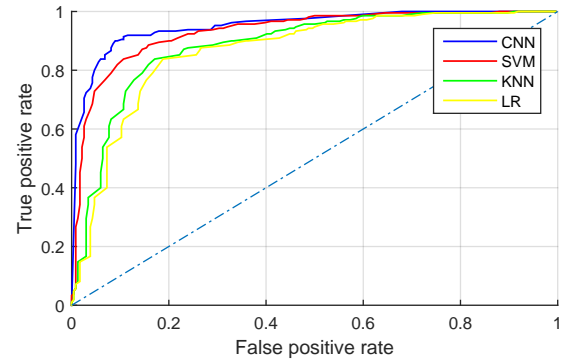

Figure 5. The ROC curves of the four tested binary face classifiers on criminality.

classifiers for the task of discriminating between criminals and non-criminals. We randomly label the faces in the very same sample set $S$ as negative and positive instances with equal probability, and redo all the above experiments of binary classification. The outcomes show that the randomly

4

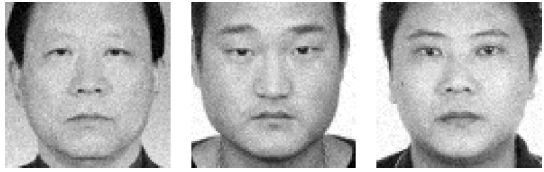

(a)

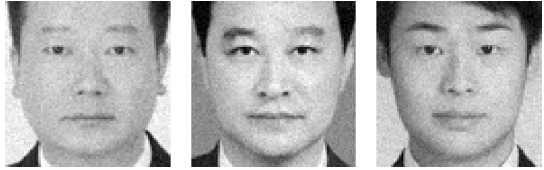

(b)

Figure 6. Noisy versions ($\sigma = 0.03$) of sample ID photos in Figure 1.

generated negative and positive instances cannot be distinguished at all; the average classification accuracy is found to be only 48%, false negative rate about 51% and false positive rate about 50%. Similar fifty-fifty observations are also made by randomly and equal-probably labeling the members of subset $S_c$ (or $S_n$) as positive and negative instances, and checking the performances of the four binary classifiers, after they are built upon and applied to the randomly shuffled classes.

In fact, we go much further along the self-critical path, and carry out the same experiments of random labeling on different sample sets of the same size (1500) and with the same variable control. Only this time the selection criteria are: case 1. standard ID photos of Chinese, female, young or middle age, no facial scars or other markings; case 2. standard ID photos of Caucasian, male, young or middle age, no facial markings; case 3. standard ID photos of Caucasian, female, young or middle age, no facial markings. In none of the three cases, any of the four classifiers manages to achieve a true positive rate higher than 53 percent on randomly labeled positive and negative instances.

The above experiments rule out that the good accuracies of the four evaluated classifiers in face-induced inference on criminality are due to data overfitting; otherwise, given the same sample size, they would also be able to distinguish between the randomly labeled positive and negative instances with significantly better chances than guessing at random. The big jump of the true positive rate from random labeling to truth labeling of the same set $S$ of sample face images can only be explained by intrinsic separability of $S_c$ and $S_n$.

As different source cameras generated the ID photos in set $S$, they might leave their signatures that, although below perception threshold in signal strength, could mislead machine learning. This issue was already dealt with at the end of Section 2 and in Section 3.1 (the use of landmark points

| noise $\sigma$ | | 0 | 0.01 | 0.03 |
|---|---|---|---|---|
| Accuracy | SVM | 88.98% | 81.03% | 73.11% |
| | CNN | 89.51% | 82.65% | 76.88% |

Table 2. The accuracies of the CNN and SVM (using combined features) face classifiers for different noise levels.

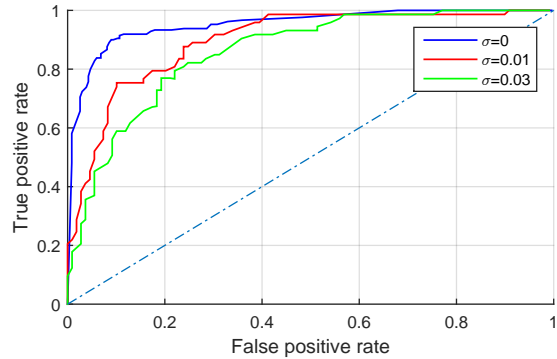

Figure 7. The ROC curves of the CNN face classifier for different noise levels.

| noise $\sigma$ | | 0 | 0.01 | 0.03 |
|---|---|---|---|---|
| Accuracy | KNN | 79.16% | 78.03% | 76.81% |
| | LR | 83.71% | 80.19% | 77.45% |
| | SVM | 82.99% | 81.52% | 79.31% |
| False alarm | KNN | 14.79% | 15.91% | 13.97% |
| | LR | 10.65% | 14.01% | 13.39% |
| | SVM | 11.57% | 14.21% | 12.51% |
| Missing | KNN | 31.48% | 31.52% | 38.53% |
| | LR | 26.20% | 30.20% | 37.41% |
| | SVM | 26.58% | 28.47% | 34.29% |

Table 3. The accuracy, false alarm and missing rates of the KNN, LR and SVM classifiers when they use landmark points as features and operate on noisy face images.

as features). To make our case even tighter we take extra steps to make the face classifiers completely immune from any potential biases of source cameras. We redo the experiments in Section 3.1 after adding Gaussian random noises of zero mean and variance $\sigma^2$ to the ID photos in $S$. The added noises will overpower any camera-dependent signatures. The performances of the classifiers on noisy input face images are reported in Tables 2 and 3, and Figure 7.

Table 2 shows that the accuracies of the CNN and SVM (using combined features) face classifiers decrease as noise level $\sigma$ increases, but they are still above 73% for SVM and 76% for CNN even when the noises become visible ($\sigma = 0.03$, see Figure 6). The ROC curves of the CNN face classifier for different $\sigma$ are plotted in Figure 7. As shown, the CNN classifier can withstand severe noise corruption

5

and still discriminate between criminals and non-criminals with high true positive rate and relatively low false positive rate.

As argued in Section 3.1, landmark points are features that are invariant to source cameras. Table 3 tabulates the accuracy, false alarm and missing rates of the KNN, LR and SVM classifiers when they use landmark points as features and operate on noisy face images. As we expected, there are no material changes in all performance numbers over different noise levels.

Also, we have tested whether the classification results are robust against slight variations in lighting environment and face orientation. We recruit 10 male Chinese students and casually take four face photographs for each of them in different environments and feed these photos to the four classifiers built on $S_c$ and $S_n$. The classification results turn out to be consistent with a probability higher than 83 per cent.

## 4. Discriminating Features

Having obtained the above strong empirical evidences for the validity of automated face-induced inference on criminality, one cannot resist the following intriguing question: what features of a human face betray its owner's propensity for crimes? In fact, the same question has captivated professionals (e.g., psychologists, sociologists, criminologists) and amateurs alike, cross all cultures, and for as long as there are notions of law and crime. Intuitive speculations are abundant both in writing [16, 12] and folklore. In this section, we try to answer the question in the most mechanical and scientific way allowed by the available tools and data. The approach is to let a machine learning method explore the data and reveal the most discriminating facial features that tell apart criminals and non-criminals. We apply the Feature Generating Machine (FGM) of Tan *et al.* [29] to the task; it identifies the red-marked regions in Figure 8 (a) as the most critical parts for the separation of criminals and non-criminals. Guided by FGM, we discover that the following three structural measurements in the critical areas around eye corners, mouth and philtrum that have significantly different distributions for the two populations in $S_c$ and $S_n$: the curvature of upper lip denoted by $\rho$; the distance between two eye inner corners denoted by $d$; and the angle enclosed by rays from the nose tip to the two corners of the mouth denoted by $\theta$. The three discriminating structural features $\rho$, $d$ and $\theta$ are shown in Figure 8 (b). We stress that the upper lip curvature $\rho$ is measured on standard ID photos taken with the person in neutral facial expression.

Let random variables $x_c$ and $x_n$ be any of the above three measurements for criminals and non-criminals. We examine the two histograms $P(x_c)$ and $P(x_n)$ and find they have rather large Hellinger distance [7]. The Hellinger distance

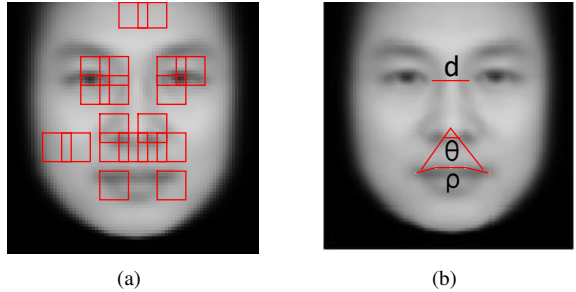

Figure 8. (a) FGM results; (b) Three discriminative features $\rho$, $d$ and $\theta$.

|  | Mean | | Variance | |
| --- | --- | --- | --- | --- |
|  | criminal | non-criminal | criminal | non-criminal |
| $\rho$ | 0.5809 | 0.4855 | 0.0245 | 0.0187 |
| $d$ | 0.3887 | 0.4118 | 0.0202 | 0.0144 |
| $\theta$ | 0.2955 | 0.3860 | 0.0185 | 0.0130 |

Table 4. The mean and variance for three normalized discriminative features $\rho$, $d$ and $\theta$.

between two probability distributions is defined as

$$H(P(x_c), P(x_n)) = \sqrt{1 - \sum_i \sqrt{P(x_c(i)) \cdot P(x_n(i))}} \quad (1)$$

Hellinger distance $H(\cdot, \cdot)$ ranges from 0 to 1, where 0 means that two probability distributions are identical and 1 means that probability distributions are different completely. In our case, the Hellinger distances are quite significant, being 0.3208, 0.2971, 0.3855 for $\rho$, $d$ and $\theta$, respectively.

In Figure 9, we plot the histograms of $\rho$, $d$ and $\theta$ after normalizing every measurement to range $[0, 1]$. The mean and variance of $\rho$, $d$ and $\theta$ for criminals and non-criminals are tabulated in Table 4. One can see in Figure 9 and Table 4 that the angle $\theta$ from nose tip to two mouth corners is on average 19.6% smaller for criminals than for non-criminals and has a larger variance. Also, the upper lip curvature $\rho$ is on average 23.4% larger for criminals than for non-criminals. On the other hand, the distance $d$ between two eye inner corners for criminals is slightly narrower (5.6%) than for non-criminals. Interestingly, in a study on perceived and measured intelligence by Chvatalova *et al.* [20], it was found that greater interpupillary distance is correlated with higher IQ for Caucasian men. But it is worth noting that the distance between two eye inner corners is a more discriminating feature than the interpupillary distance for the classification of criminals.

6

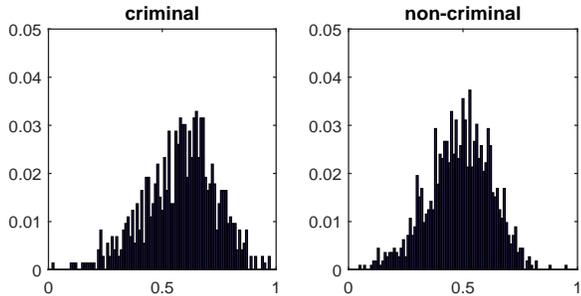

(a) Histograms of $\rho$

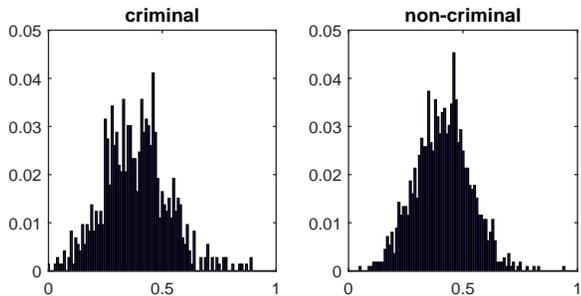

(b) Histograms of $d$

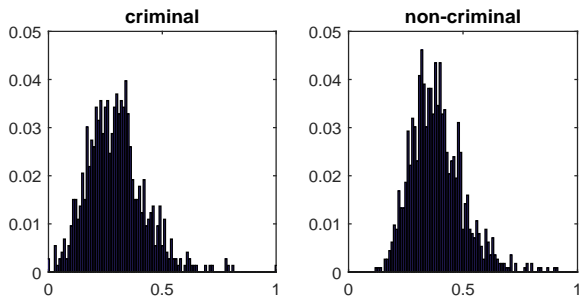

(c) Histograms of $\theta$

Figure 9. Histograms of the three discriminative features.

## 5. Face Clustering on Manifolds

Encouraged by the newly found success of machine learning in face-induced inference on criminality and the knowledge of the underlying discriminative features, the next question is logically, what are the typical (average) faces for criminals and non-criminals? Indeed, some researchers embarked on this type of enquiry and published on topics such as the characteristic face of female beauty [25], distinctive faces of certain races [22] and nationalities (e.g., Japanese vs. Koreans vs. Chinese) [38]. Previous authors attempted to find a typical face (i.e., an average of some sort) for a category of people, such as Chinese, beautiful female, criminals, and the alike. But in this section, we argue against the practice of using a single representative face for a social label such as beauty and criminal, even when the labeling is highly consensual.

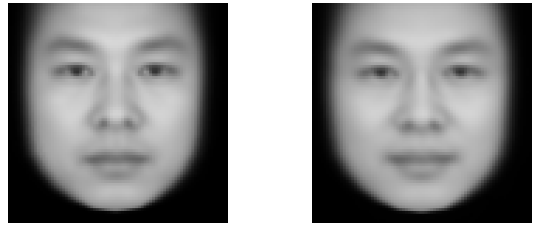

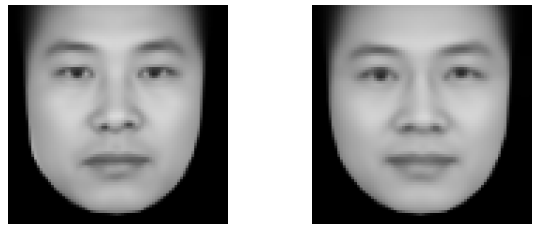

Figure 10. (a) and (b) are "average" faces for criminals and non-criminals generated by averaging of eigenface representations ; (c) and (d) are "average" faces for criminals and non-criminals generated by averaging of landmark points and image warping.

Although the antithesis of criminals and non-criminals is very strong, conventionally-defined average faces of the two populations $S_c$ and $S_n$ appear hardly distinguishable as demonstrated in Figure 10. The displayed average faces are generated either by averaging of landmark points and image warping [15], or by averaging eigenface representations [36] over the sample set. In order to understand and solve the puzzle of failing to find an average face of criminals that is sufficiently distinctive from that of non-criminals, we look into the high-dimensional distributions of $S_c$ and $S_n$, and the structural statistical relationships between the two populations in $S_c$ and $S_n$.

The seeming paradox that $S_c$ and $S_n$ can be classified but the average faces of $S_c$ and $S_n$ appear almost the same can be explained, if the data distributions of $S_c$ and $S_n$ are heavily mingled and yet separable by a complex surface discriminant. This thought leads to the suspicion: perhaps, manifolds are good representations for the data clouds of $S_c$ and $S_n$; in other words, faces of criminals and non-criminals are assumed to populate two quite distinctive manifolds. Our manifold hypothesis is also promoted by observing that the differences between faces can be modeled as the results of continuous morphing.

To test this hypothesis, we compute the cross-class average manifold distance $D_\times$ between two subsets $S_c$ and $S_n$, and the in-class average manifold distances $D_c$ and $D_n$ for $S_c$ and $S_n$, and compare them. Let $d(f_i, f_j)$ denote the geodesic distance between two different faces $f_i$ and $f_j$, which is defined as the length of shortest the path between $f_i$ and $f_j$ in the sparse neighborhood graph [30]. The

7

aforementioned average manifold distances are defined as

$$D_\times = \frac{1}{|S_c||S_n|} \sum_{f_i \in S_c, f_j \in S_n} d(f_i, f_j)$$
$$D_c = \frac{2}{|S_c||S_c - 1|} \sum_{f_i, f_j \in S_c; i \neq j} d(f_i, f_j) \quad (2)$$
$$D_n = \frac{2}{|S_n||S_n - 1|} \sum_{f_i, f_j \in S_n; i \neq j} d(f_i, f_j)$$

The sample sets of the criminal and non-criminal face images reveal that $D_c > D_\times > D_n$. These inequalities of manifold distances and the fact that $S_c$ and $S_n$ have almost the same sample mean vectors suggest the possibility that the two manifolds of $S_c$ and $S_n$ are concentric, with the non-criminal manifold lying in the kernel with a smaller span and the criminal manifold forming an outer layer of large spread.

To verify our intuition, let us visualize the sample distributions of $S_c$ and $S_n$ after a dimensionality reduction processing. We adopt the nonlinear dimensionality reduction method called Isomap [30] to compute a quasi-isometric, low-dimensional embedding of our ultra-high-dimensional data sets. Isomap uses the geodesic distance $d(a, b)$ between two points $a$ and $b$ on a manifold, which is defined to be the sum of edge weights along the shortest path connecting $a$ and $b$ in the sparse neighborhood graph. The largest $n$ eigenvectors of the geodesic distance matrix represent the coordinates in the new $n$-dimensional Euclidean space.

Figure 11 depicts the relationship of residual variance and Isomap dimensionality; it indicates that the original ultra-high dimensional data set can be represented reasonably well in a subspace of four to six dimensions. The data clouds of criminals and non-criminals can be visualized in Figure 12, in which the first four most significant dimensions of the Isomap are represented by the $x$, $y$, $z$ axis, and the color temperature.

All above analyses and visualization indicate that a subjectively meaningful typical face of criminals, even after controlled for race, gender, nationality and age, simply does not exist. Instead, there should be few mutually distinctive subtypes of criminal faces. In other words, the correct answers lie in data clustering on manifolds. We use the geodesic K-means clustering [4] to discover these representative subtypes in sample sets $S_c$ and $S_n$. In the geodesic $K$-means clustering of $S_c$, first $K$ faces are randomly chosen as tentative centroids on the manifold of $S_c$, denoted by $g_k$, $1 \leq k \leq K$. Then each face $f_i \in S_c$ is assigned to cluster $C_j$ whose centroid is the closest to $f_i$ in the geodesic distance $d$,

$$j = \arg\min_{1 \leq k \leq K} d(f_i, g_k). \quad (3)$$

Next the cluster centroids are updated by the nearest neigh-

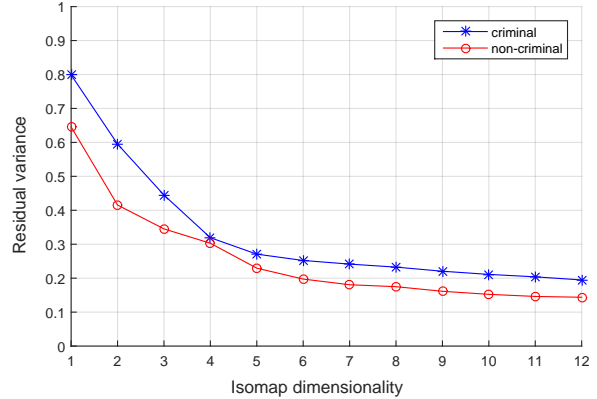

Figure 11. The relationship of residual variance and Isomap dimensionality.

bor rule:

$$g_k = \arg\min_{f_i \in C_k} \sum_{f_j \in C_k} d(f_i, f_j). \quad (4)$$

The above steps iterate till the cluster centroids converge. For final cluster $C_k$, $1 \leq k \leq K$, we compute the representative face of subtype $k$ of criminals, by warping all faces $f_i \in C_k$ so that the landmark points of $f_i$ are aligned with corresponding ones of $g_k$, and averaging all warped faces.

The same clustering and averaging processes are also carried out on $S_n$ to produce subtype faces for non-criminals.

Figure 13 displays four subtypes of criminal faces in $S_c$ and three subtypes of non-criminal faces in $S_n$ that are synthesized by the above described processes. These computer-generated representative faces for criminals and non-criminals appear to be in agreement with the intuition of 50 Chinese students of both sexes who participated in an subjective test. In this experiment, every participant has to make a binary decision on each of these seven synthesized subtype faces (presented in random order) and assign score $-1$ for being more criminal like, or $+1$ otherwise. All participants do not know that these faces are synthesized nor how many of them are representatives of criminals. The average scores for these synthesized faces corresponding cluster centroids are shown in Figure 13.

As the $K$-means clustering is unsupervised, we need to justify why there are four subtypes of criminal faces in $S_c$ but only three subtypes of non-criminal faces in $S_n$. Figure 14 depicts how the variation within a cluster decreases in the number of clusters $K$ for both $S_c$ and $S_n$. The figure clearly illustrates that there are four well separable clusters (distinctive facial appearances) for criminal faces as the in-cluster variation drops quickly till $K = 4$, while the faces of non-criminals do not form as many separable clusters in geodesic distance on the manifold of $S_n$.

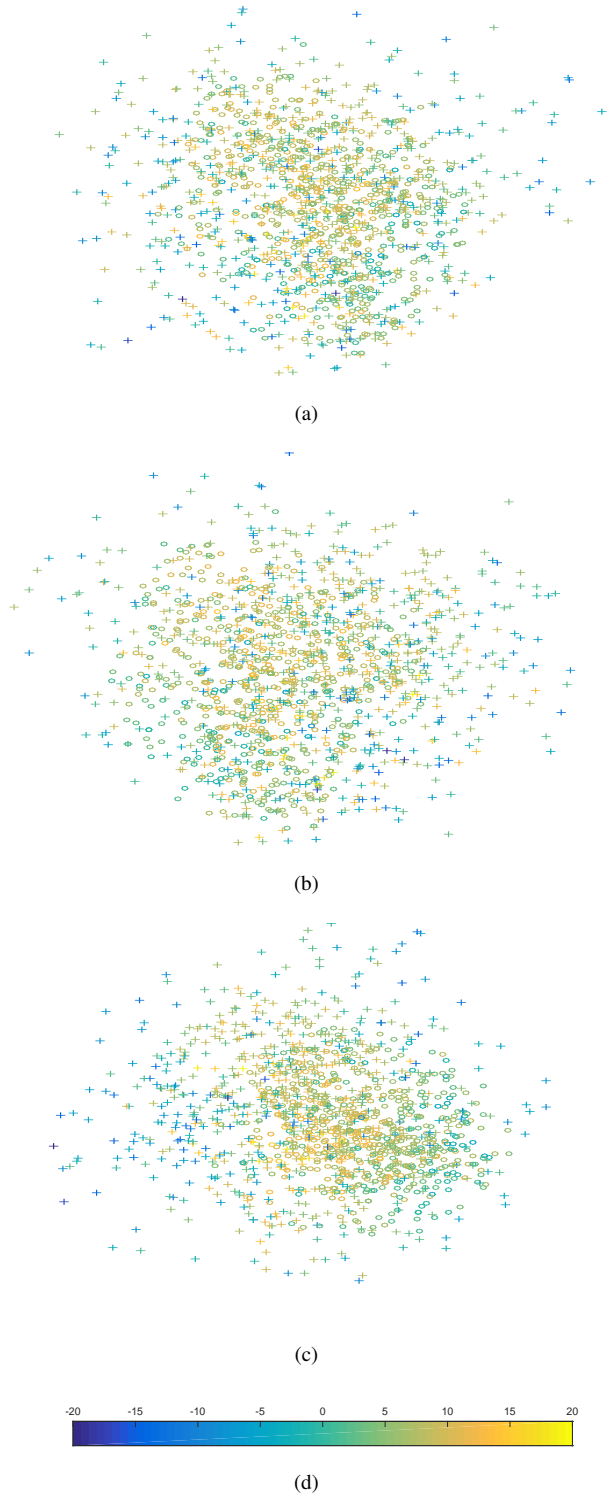

(a)

(b)

(c)

(d)

Figure 12. Data clouds ("+" labels criminals and "o" labels non-criminals) in the first four most significant dimensions of the Isomap, viewed in three different perspectives.

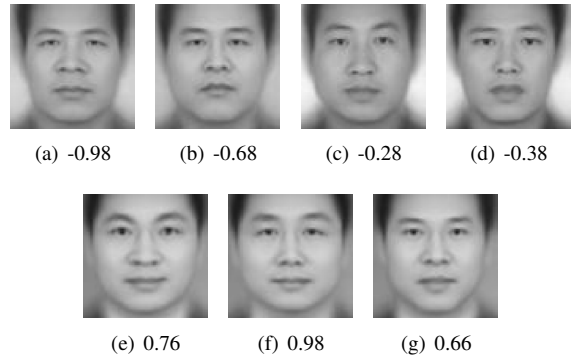

(a) -0.98   (b) -0.68   (c) -0.28   (d) -0.38

(e) 0.76   (f) 0.98   (g) 0.66

Figure 13. (a), (b), (c) and (d) are the four subtypes of criminal faces corresponding to four cluster centroids on the manifold of $S_c$; (e), (f) and (g) are the three subtypes of non-criminal faces corresponding to three cluster centroids on the manifold of $S_n$. The number associated with each face is the average score of human judges (-1 for criminals; 1 for non-criminals).

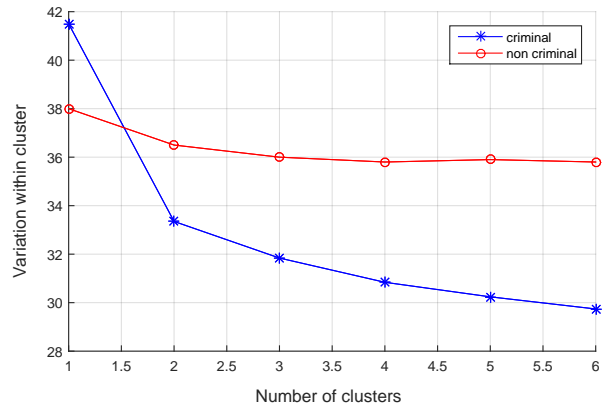

Figure 14. The relationship of variation within a cluster and number of clusters for criminal and non-criminal data set.

The above data analysis and visualization allow us to make the following interesting and meaningful conclusion, which holds at least for the class of human subjects being examined here, i.e., male Chinese of young or middle age. Although criminals are a small minority in total population, they have appreciably greater variations in facial appearance than general public. This coincides with the fact that all law-biding citizens share many common social attributes, whereas criminals tend to have very different characteristics and circumstances, some of which are quite unique of the individual's own.

## 6. Conclusions

We are the first to study automated face-induced inference on criminality free of any biases of subjective judgments of human observers. By extensive experiments and vigorous cross validations, we have demonstrated that via

9

supervised machine learning, data-driven face classifiers are able to make reliable inference on criminality. Furthermore, we have discovered that a law of normality for faces of non-criminals. After controlled for race, gender and age, the general law-biding public have facial appearances that vary in a significantly lesser degree than criminals.



\end{document}